# CNN based Multi-Instance Multi-Task Learning for Syndrome Differentiation of Diabetic Patients

Zeyuan Wang
*School of Information Technologies*
*The University of Sydney*
Sydney, Australia
zwan7221@uni.sydney.edu.au

Josiah Poon
*School of Information Technologies*
*The University of Sydney*
Sydney, Australia
josiah.poon@sydney.edu.au

Shiding Sun
*School of Mathematics*
*Renmin University of China*
Beijing, China
sunshiding@ruc.edu.cn

Simon Poon
*School of Information* Technologies
*The University of Sydney*
Sydney, Australia
simon.poon@sydney.edu.au

*Abstract*—Syndrome differentiation in Traditional Chinese Medicine (TCM) is the process of understanding and reasoning body condition, which is the essential step and premise of effective treatments. However, due to its complexity and lack of standardization, it is challenging to achieve. In this study, we consider each patient's record as a one-dimensional image and symptoms as pixels, in which missing and negative values are represented by zero pixels, labeled by one or more syndromes in diabetes. The objective is to find relevant symptoms first and then map them to proper syndromes, that is similar to the object detection problem in computer vision. Inspired from it, we employ multi-instance multi-task learning combined with the convolutional neural network (MIMT-CNN) for syndrome differentiation, which takes region proposals as input and output image labels directly. The neural network consists of region proposals generation, convolutional layer, fully connected layer, and max pooling (multi-instance pooling) layer followed by the sigmoid function in each syndrome prediction task for image representation learning and final results generation. On the diabetes dataset, it performs better than all other baseline methods. Moreover, it shows stability and reliability to generate results, even on the dataset with small sample size, a large number of missing values and noises.

## I. Introduction

Chronic complications of diabetes, including visual impairment, kidney disease, neuropathy, peripheral vascular disease, heart disease, and stroke, are the most important causes of disability and death in diabetic patients. It becomes a sever public health problem in recent years [1]. Because Tradition Chinese Medicine (TCM) does not treat certain symptoms and signs as isolated, but treat the patient as a whole, it has the great advantages in the treatment of diabetes and its complications. However, due to the current lack of uniform, standardized and objective criteria for the syndrome differentiation in TCM [2], the clinical efficacy of preventing and treating diabetes is not convincing. Moreover, because of the many-to-many relationship between symptoms and syndromes, it's difficult to conduct quantitative syndrome differentiation. Therefore, exploring their relations and developing methods for classification or prediction tasks, have attracted more and more TCM researchers' attention [3].

Latent structure analysis [4-9], has become one of the most successful unsupervised learning methods for TCM syndrome differentiation. It provides standardization and objectiveness for syndrome differentiation when there is no reference standard. For quantitative syndrome differentiation, some innovative classification methods have been proposed. Liu et al. [10] applied feature selection combined with the ML-KNN and successfully explored the relations between symptoms and syndromes for the patients with chronic gastritis. With the same dataset, they developed the deep belief network [11] for the same purpose. The method treats syndromes as joint prediction tasks and successfully takes the relationship among symptoms into consideration which is consistent with the TCM diagnosis process. Wang et al. [12] mainly focused on the syndrome differentiation for patients with liver cirrhosis. They proposed a feature selection first and accuracy-weighted majority voting second method based on the data from both TCM view and western medicine view. Wang et al. [13] creatively proposed a method for selecting the true labels first and conducting multiple tasks prediction at last for chronic fatigue patients. It not only shows outstanding performance but also provides confidence evaluation for syndrome differentiation. Zhao et al. [3] were first bringing the idea of multi-instance learning (MIL), proposed by Dietterich et al. [14, 15] for drug molecule activity prediction, to this field. They developed a novel method, MRS-MIL, for patients with AIDS. In MIL, it takes a series of labeled bags, each of which contains many unlabeled instances, as input to train the classifier. Patients can be considered as bags and each contained instance is one of or part of input symptoms. It has the ability to discover the key instance in the bag, in other words, the key symptoms for each syndrome. Also, it is robust to the unbalance data and small sample size [3], which are two of main problems in analyzing clinical data. Their work shows that multi-instance learning is quite effective and suitable for resolving the syndrome differentiation problem.

These algorithms are all treating syndrome differentiation problem as multiple tasks prediction, since in most cases, patients have more than one syndrome at the same time. And multi-instance learning shows its ability for clinical prediction tasks. Therefore, in our proposed method, we took both ideas combined with the convolutional neural network (CNN). Because MIL only works for the single-label scenarios, it requires multiple tasks problem to be resolved by applying MIL for each task once [16]. About CNN, it is a widely used artificial neural network in computer vision classification tasks. And it is inspired by the study of Hubel et al. [17] that the transmission of visual information from the retina to the brain is triggered by multiple levels of receptive fields.

The reason we applied the algorithm in the computer vision field is based on a simple but very effective idea, treating each patient as an image (bag) and generating labels based on the key region proposals (instances) of the image.

When analyzing TCM data, except for small size of samples and unbalanced binary labels, there still are two main issues affecting the performance of models: a large number of negative values and irrelevant features, i.e. noise. As for many irrelevant features, if we consider patients as images, it can be translated as detecting and classifying a very small object in the image, a classic object detection problem in computer vision. About a large number of negative values in each feature, they represent two meanings: (1) the patients didn't take this test, in other word, missing value; (2) the patients did take this test but didn't have this symptom. It's almost impossible to distinguish these two situations only based on the dataset and difficult to find the appropriate algorithms to make the final classification. However, if we consider patients' records as images and use zero pixels to represent missing values and negative values, these two situations can be interpreted as, in this image, these pixels are missing or originally black here (zero pixels in the image denotes to the black color). And CNN is robust to these two situations when classifying images.

Based on these ideas, we develop our method and our experiment results demonstrate that the problems in analyzing the TCM data are all be resolved perfectly.

## II. MATERIALS AND METHODS

### A. Dataset of Diabetes in TCM

The diabetes dataset comes from the Guang'anmen Hospital, China Academy of Chinese Medical Sciences, Beijing. It has been analyzed for significant herb interactions discovery [18] and symptom-herb patterns [19], exploring the relationship among symptoms and herbs, to support clinical decision-making process.

In the dataset, there are 1915 outpatient records consisting of four parts: symptoms, syndromes, herbal prescriptions and outcomes which are annotated as effective or ineffective of herbal treatments by TCM clinical doctors based on the changes of blood glucose and hemoglobin levels. There are 1180 records selected for the model development, based on the following inclusion criteria:

1) outpatient record completed;
2) the frequency of the syndrome larger than 50;
3) at least two symptoms existing.

In them, there are 186 different symptoms from TCM four diagnosis and 12 main syndromes:

1) qi and yin deficiency;
2) collaterals stasis;
3) hyperactivity of liver-yang;
4) fire excess from yin deficiency;
5) the stagnation of liver qi and stomach heat;
6) yin deficiency with internal heat;
7) blood clots hinder vessels;
8) imbalance between heart-yang and kidney-yin;
9) combination of phlegm and heat;
10) spleen deficiency with stomach heat;
11) block of collateral;
12) meridian obstruction.

In addition, for each patient, there are at most 18 symptoms and 4 syndromes at the same time. The dataset is in one-hot format and all missing values are replaced by the negative value, 0. So, there are a large number of negative values in this dataset, which is challenging to conduct the prediction tasks.

### B. Convolutional Neural Network (CNN)

CNN takes images as input and converts each of them into a series of feature maps through alternately connected convolutional and pooling layers. At last, through fully connected layer, output one or more labels of the image. The whole network can be supervised and trained through back propagation [20].

There are four basic units in CNN [21]: convolutional layer, activation function, pooling layer and fully connected layer. For the convolutional layer, which is considered as the feature extractor, it can be formulated as:

$$f_c(x) = act(\sum_{i,j}^{n} W_{c(n-i)(n-j)} x_{ij} + b_c) \quad (1)$$

where $W_c$ and $n$ represent the weight matrix and size of perception field. $b_c$ denotes the bias vector and $act(x)$ is the activation function which makes the neural network able to approximate any nonlinear function. In our proposed model, we applied activation function ReLU:

$$f_r(x) = max(0, x) \quad (2)$$

For the pooling layer, there are many types of them. It figures out a value that represents the local feature, which greatly reduces the parameters. Common are max pooling, mean pooling and L2 pooling. We chose max pooling and it is done by applying the max operation to non-overlapping sub-regions of the input. The size of sub-regions is pre-defined.

About the fully connected layer, which can be considered as the linear projection layer, it projects features extracted from the former layers to the linear space.

At last, sigmoid or softmax function is implemented for the final classification. As our prediction tasks are all binary, we apply the sigmoid function for each task which is formulated as:

$$f_s(x) = 1/(1 + e^{-x}) \quad (3)$$

### C. Multi-Instance Learning

In common supervised learning, an input sample $x_i \in X$ is represented by an instance, i.e. feature vector, and labeled by $y_i \in Y$, where $X, Y$ denote the input sample set and label set. The task is to learn a function, mapping $X$ to $Y$ [22].

In multi-instance learning, each sample $x_i$ is a bag, which contains several unlabeled instances $x_{ij} \in \mathbb{R}^{d \times 1}$, with a binary label $y_i$. The task is to, for each bag, take contained instances as input and decide whether they are positive or not. If there is at least one positive instance in the bag, the bag is labeled positive, otherwise, negative. It is formulated as:

$$y_i = \begin{cases} 0, & all\ y_{ij} = 0 \\ 1, & otherwise \end{cases} \quad (4)$$

where $y_{ij}$ is the label of the $j^{th}$ instance in the $i^{th}$ bag.

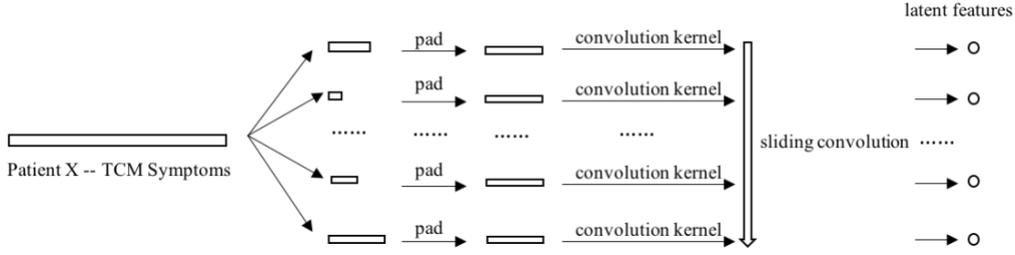

Figure 3: The process of instance representation

Since labels of instances are not given, it is challenging to conduct training process. However, it is easy to be solved if combine with the neural network and end-to-end training. Cause if re-formulate the assumptions of multi-instance learning problem, we can find that:

$$y_i = \max_j \{y_{ij}\} \quad (5)$$

which can be replaced by the max pooling layer in neural network and it is differentiable, so it can be updated and trained through back propagation. The process is also known as multi-instance pooling for selecting the most likely positive instance as the bag representation [23].

The whole process can be considered as key information detection first and then make the final decision based on it. If we consider patients as bags (images) and symptoms as instances (region proposals), it is consistent to the TCM diagnosis process. Because in practice, patients won't take all clinical examination and they are diagnosed only based on the key symptoms or signs in existing records.

Also, in most cases, the number of negative samples is much larger than the positive samples. And multi-instance learning generates results only based on the key instances, so it won't be highly influenced by unbalanced samples. For now, it has been widely used in the bioinformatics researches.

### D. Multi-Task Learning

Compared to the traditional single task learning, multi-task learning is a method to solve multiple related tasks at the same time. It aims to improve the performance of each prediction task by learning them jointly through shared representation [24]. A simple example is shown in Figure 1 and the process can be defined as:

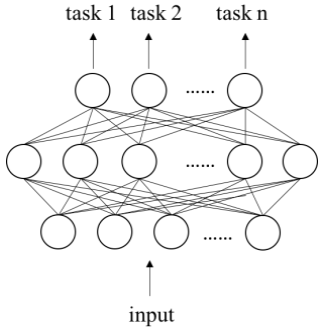

Figure 1: An example of Multi-Task Learning

Given the input sample $x_i \in X$ labeled by $Y_i \in Y$, where $X, Y$ denote the input sample set and label set, and $Y_i$ is a set with $l$ labels $\{y_{i1}, y_{i2}, ..., y_{il}\}$. Through training, learn a function, mapping $X$ to $Y$.

### E. CNN based Multi-Instance Multi-Task Learning (MIMT-CNN)

This section lays out the proposed model, MIMT-CNN, in which region proposals generation is the first step. Inspired from the classification algorithm Random Forest [25] and object detection algorithm Fast R-CNN [26], for each patient, we randomly select features in random size several times as region proposals (instances). The process is similar to generate region boxes on the image and try to decide which boxes contain the object and what's the label or category of the object. The regional proposal generation times and the maximum size of them can be adjusted depends on the clinical needs.

Subsequently, since the input of convolutional layer should be in the same size, we pad each region proposals (instances) to the maximum size. The process is shown as Figure 2.

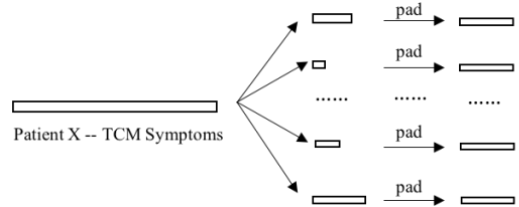

Figure 2: The process of region proposals (instances) generation

For each generated region proposal, convolution operation followed by activation function ReLU is implemented, projecting it to a single value which is the latent feature representing this region proposal (instance). When applying convolution kernel, it slides on the region proposals one by one. The process is shown as Figure 3.

Then, we apply fully connected layer and max pooling (multi-instance pooling) layer to select the most likely positive region proposal (instance) for each task as the image (bag) representation. It means that for each syndrome, only select the most important symptoms to support the differentiation. Finally, for each syndrome, the sigmoid function and cross entropy are applied for the binary classification and loss calculation. All independent losses are summed up as the final

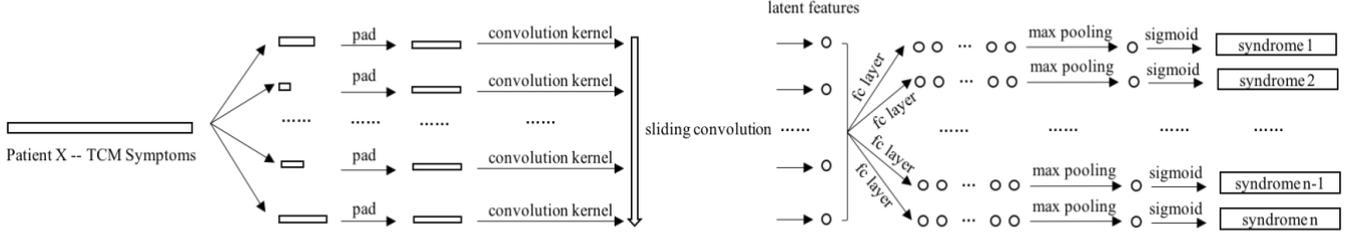

Figure 4: The model architecture

loss to be optimized through back propagation. The final loss function is formulated as following:

$$loss(y, z) = -\sum_{i=1}^{m} \sum_{j=1}^{n} (y_i^j \log z_i^j + (1 - z_i^j) \log(1 - z_i^j)) \quad (6)$$

where $z = \{z_i^j\}_{i,j=1}^{m,n}$ denotes output matrix and $y = \{y_i^j\}_{i,j=1}^{m,n}$ represents the labels of tasks. The whole process is shown as Figure 4.

### III. EXPERIMENTS

All experiments were run in 5-fold cross validation. For testing the influence of region proposals, we tried the maximum size of them in 5, 10, 15, 20 and 25 and the generation times in 100, 500, 1000, 1500 and 2000. In fully connected layer, hidden units were set to 64. The neural network was trained in 100 epochs with batch gradient descent and Adam optimizer. The learning rate was set as 0.1.

Our method has the ability to discover the key information for syndrome differentiation, so it is robust to noises and small dataset. For testing the influence of small sample size, we randomly selected 90%, 80%, 70% and 60% samples from training data in each fold to develop the model. For testing the reliability when much noise existing, we added 5, 10, 15 and 20 new features with randomly generated binary values, to the feature set. These new features are served as noises and they are totally meaningless in practice.

The following four metrics were implemented to measure the performance of proposed and baseline models [27].

- Mean Average Precision: The mean of the average precision scores for tasks. And average precision score is the area under Precision-Recall curve. The value is between 0 and 1 and higher is better.

$$MAP(f) = \frac{1}{n} \sum_{i=1}^{n} AveP(task_i) \quad (7)$$

- Coverage: Compute how far on average to go through the list of labels to cover the true labels. The value is between 0 and the smaller the value, the better the model performance.

$$coverage(f) = \frac{1}{p} \sum_{i=1}^{p} \max_{y_j \in y^i} (1 - f(x_i, y)) - 1 \quad (8)$$

- Subset Accuracy: Calculate the proportion of samples that the predicted lables are exactly the same as true labels. The value is between 0 and 1 and the larger the value, the better the model performance.

$$Subset\ Accuracy(f) = \frac{1}{p} \sum_{i=1}^{p} 1\{f(x^i) = y^i\} \quad (9)$$

- Hamming Loss: The degree of inconsistency between the predicted label and the true label of the sample. The value is between 0 and 1. The smaller the value, the better the model performance.

$$Hamming\ Loss = \frac{1}{p} \sum_{i=1}^{p} \frac{1}{q} |h(x^i) \Delta y^i| \quad (10)$$

where $\Delta$ represents the symmetry difference of the predicted label set and true label set.

### IV. RESULTS AND DISCUSSION

#### A. Comparison of different maximum size of region proposals (instances).

In clinical, doctors diagnose only based on the key symptoms. However, as for how many key symptoms are enough to support diagnosis, every clinical doctor has his own opinion. In diabetes dataset, for testing this, we set the region proposal generation times to 500. The maximum size was tried as 5, 10, 15 and 20, which meant either of 5, 10, 15 or 20 symptoms were enough to support syndrome differentiation. To evaluate the model performance, we applied mean average precision.

As shown in Figure 5, as the maximum size increases, it brings more and more information, and the model performance is better and better. However, after 10, the model performs worse, since it brings more and more noise and correlations among symptoms.

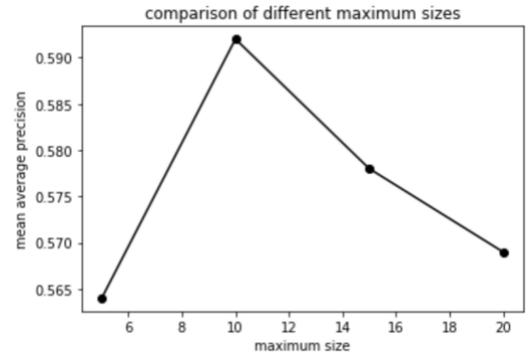

Figure 5: The comparison among different maximum sizes of region proposals

#### B. Comparison of different region propsal generation times

We selected the maximum size with the best model performance, 10. The region proposal generation times were set to 100, 500, 1000, 1500 and 2000 for comparison to decide how many generation times were the best. The mean average precision was used to measure the model performance. As the Figure 6 shown, 500 times generation are enough to cover all necessary information for the final classification. Therefore, we set the generation times as 500 for the future experiments.

TABLE I. RESULTS OF DIFFERENT MODELS (IN%)

|          | Mean Average Precision | Coverage | Subset Accuracy | Hamming Loss |
|----------|------------------------|----------|-----------------|--------------|
| ML-KNN   | 0.463                  | 0.357    | 0.080           | 0.117        |
| Rank-SVM | 0.469                  | 0.360    | 0.109           | 0.110        |
| BSVM     | 0.490                  | 0.340    | 0.163           | 0.100        |
| MLP      | 0.558                  | 0.356    | 0.239           | 0.094        |
| DBN      | 0.539                  | 0.355    | 0.233           | 0.098        |
| REAL     | 0.488                  | 0.339    | 0.150           | 0.105        |
| MIMT-CNN | **0.573**              | **0.321**| **0.259**       | **0.081**    |

TABLE II. PRECISION AND RECALL FOR EACH SYNDROME PREDICTION TASK (IN%)

|                                              | Precision | Recall |
|----------------------------------------------|-----------|--------|
| qi and yin deficiency                        | 0.932     | 0.669  |
| collaterals stasis                           | 0.876     | 0.351  |
| hyperactivity of liver-yang                  | 0.989     | 0.353  |
| fire excess from yin deficiency              | 0.710     | 0.418  |
| the stagnation of liver qi and stomach heat  | 0.821     | 0.452  |
| yin deficiency with internal heat            | 0.805     | 0.571  |
| blood clots hinder vessels                   | 0.912     | 0.754  |
| imbalance between heart-yang and kidney-yin  | 0.993     | 0.553  |
| combination of phlegm and heat               | 0.769     | 0.370  |
| spleen deficiency with stomach heat          | 0.625     | 0.366  |
| block of collateral                          | 0.650     | 0.452  |
| meridian obstruction                         | 0.818     | 0.227  |

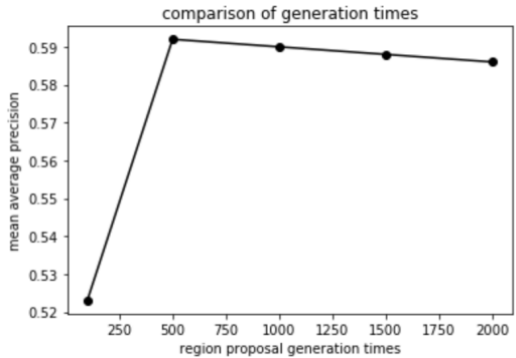

Figure 6: The comparison among different region proposal generation times

*C. Comparison of different multi-task learning models*

As the problem mainly belongs to the multi-task learning problem, we implemented following models as baselines:

- ML-KNN: We normalized all continuous variables using the extremum method and the neighbor number was set at 20. Euler distance was used as distance measurement function.
- Rank-SVM: Linear kernel function and maximum iterations were set to 50.
- BSVM: Linear kernel function.
- MLP: We chose one hidden layers with 128 neurons. Learning rate was 0.1 for 100 epochs with Adam optimizer.
- DBN: We selected one hidden layer with 256 neurons and learning rate was 0.1 for 100 epochs with Adam optimizer.
- REAL: 20 features were selected for each syndrome.
- MIMT-CNN: Maximum size of region proposals was 10 and the generation times of them was 500.

All four metrics are calculated for the compairson. As the Table 1 shown, among all methods, MIMT-CNN performs much better than the baseline models. Also, as we can see, neural networks are better than other mehtods. The possible reason is that, there are a large number of negative values and neural networks can consider them as inactivated neurons which doesn't influence other neurons receiving information too much.

*D. Performance for each task*

For deeply understanding the model performance, we applied precision and recall metrics to measure each syndrome differentiation task.

- $Precision = \frac{True\ positive}{(True\ Positive+False\ Positive)}$ (11)

- $Recall = \frac{True\ Positive}{(True\ Positive+False\ Negative)}$ (12)

As the Table 2 shown, for syndrome differentiation tasks except spleen deficiency with stomach heat and block of collateral, precision is much high. In other word, for patients if our model predicts some syndromes, in most cases, they are correct. Especially for the prediction of qi and yin deficiency, hyperactivity of liver-yang, blood clots hinder vessels, imbalance between heart-yang and kidney-yin, the precision of them are all above 0.9.

As for spleen deficiency with stomach heat and block of collateral, the model is a little bit failed to discover the key information to support diagnosis. About recall scores, they are not as good as precision scores, which is common in analyzing medical datasets. Except for the prediction tasks of qi and yin deficiency, yin deficiency with internal heat, blood clots hinder vessels and imbalance between heart-yang and kidney-yin, other recall scores are all under 0.5. In the process of model looking for key information, it cannot find the precise and complete latent features representing the patient which makes the model generate diagnosis results conservative.

*E. Influence of small sample size*

We randomly took 90%, 80%, 70% and 60% data from the training dataset in each fold for model development. For comparison, we applied ML-KNN as the baseline model. The mean average precision was applied for the performance test. As the result shown in the Figure 7, although the dataset is smaller and smaller, the MIMT-CNN doesn't change too much and the mean average precision only slightly decreases. From 100% to 60%, the mean average precision is reached at 0.573, 0.551, 0.540, 0.523 and 0.511. However, ML-KNN performs much worse than our method and reaches at 0.463, 0.410, 0.381, 0.363 and 0.340. It drops more heavily as the sample size decreases. We conclude that MIMT-CNN is still reliable even with the small sample size.

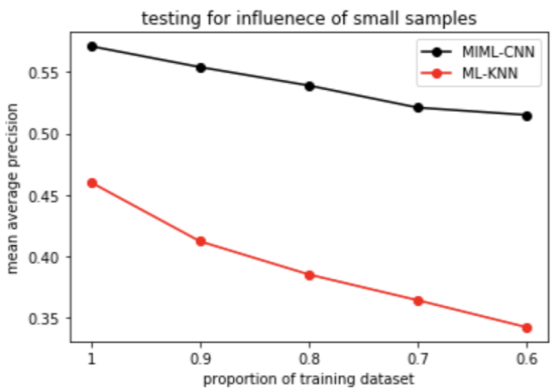

Figure 7: The test for the influence of small samples

*F. Influence of noises*

We randomly generated 5, 10, 15 and 20 binary features, which were totally meaningless in clinical and only used for the experiment. For comparison, ML-KNN was implemented again as the baseline model. The mean average precision was used for the performance measurement.

As the Figure 8 shown, the mean average precision of MIMT-CNN is 0.560, 0.557, 0.536 and 0.532 separately and if we don't add noise features, the mean average precision is 0.573. As for ML-KNN, it reaches at 0.439, 0.414, 0.389 and 0.385. Compared to ML-KNN, our method performs more steadily even with a large number of noise values.

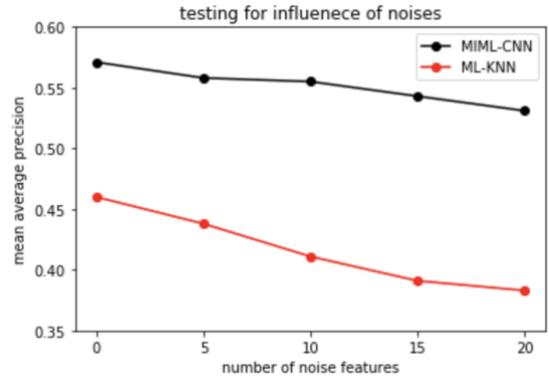

Figure 8: The test for the robustness to a large number of noise values

## V. CONCLUSION

To perfectly imitate the TCM diagnosis process and explore the relationship between symtpoms and syndromes, the CNN based multi-instance multi-task learning method (MIMT-CNN) is presented in this paper. The study shows its outstanding performance for information expression and knowledge discovery in syndrome differentiation.

Two experiments are set up for determining maximum size and generation times of region proposals. The results indicate that, for diabetic patientis, up to 10 symptoms support syndrome differentiation and 500 generation times are enough to cover all necessary symptoms information. Moreover, our method is compared with baseline models. The results demenstrate that it outperforms all baseline models on all measurement metrics.

There also are two experiments set up for testing the robustness to small sample size and a large number of noise values. The results reveal that our method still can generate the reliable outputs even in these two situations.

The study shows that MIMT-CNN is very effective to deal with TCM syndrome differentiation problems. It also can support the establishment of diagnosis criteria and supply the guide for clinical practice.